\documentclass[letterpaper, 10 pt, conference]{IEEEconf}
\IEEEoverridecommandlockouts      

\usepackage{siunitx}
\usepackage{times}
\usepackage{graphicx}
\usepackage{svg}
\usepackage[normalem]{ulem}
\usepackage{array}
\graphicspath{{images/}}
\usepackage{subcaption}
\usepackage{tikz}

\title{\LARGE \bf An Experimental Setup to Test Obstacle-dealing Capabilities of Prosthetic Feet}

\author{Anna Pace\authorrefmark{4}$^{1}$, Lukas Proksch$^{2}$, Giorgio Grioli$^{1,3}$, Oskar C. Aszmann$^{2}$, Antonio Bicchi$^{1,3,4}$, Manuel G. Catalano$^{1}$ 
\thanks{*This work was supported by the European Research Council Synergy Grant Natural BionicS (NBS) project (Grant Agreement No. 810346).}
\thanks{$^{1}$with the Soft Robotics for Human Cooperation and Rehabilitation Lab, Fondazione Istituto Italiano di Tecnologia, 16163 Genoa, Italy.}%
\thanks{$^{2}$with the Clinical Laboratory for Bionic Extremity Reconstruction, Division of Plastic and Reconstructive Surgery, Medical University of Vienna, Austria.}%
\thanks{$^{3}$with Centro di Ricerca “Enrico Piaggio”, University of Pisa, 56122 Pisa, Italy.}
\thanks{$^{4}$with Department of Information Engineering, University of Pisa, 56122 Pisa, Italy.}%
\thanks{\authorrefmark{4} Corresponding author: \tt\small anna.pace@iit.it}
}

\newcommand\copyrighttext{%
  \footnotesize \textcopyright\ \the\year{} IEEE. Personal use of this material is permitted.  Permission from IEEE must be obtained for all other uses, in any current or future media, including reprinting/republishing this material for advertising or promotional purposes, creating new collective works, for resale or redistribution to servers or lists, or reuse of any copyrighted component of this work in other works.}

\newcommand\copyrightnotice{%
\begin{tikzpicture}[remember picture,overlay]
\node[anchor=south,yshift=10pt] at (current page.south) {\fbox{\parbox{\dimexpr\textwidth-\fboxsep-\fboxrule\relax}{\copyrighttext}}};
\end{tikzpicture}%
}

\begin{document}

\maketitle

\begin{abstract}
Small obstacles on the ground often lead to a fall when caught with commercial prosthetic feet. Despite some recently developed feet can actively control the ankle angle, for instance over slopes, their flat and rigid sole remains a cause of instability on uneven grounds. Soft robotic feet were recently proposed to tackle that issue; however, they lack consistent experimental validation. Therefore, this paper describes the experimental setup realized to test soft and rigid prosthetic feet with lower-limb prosthetic users. It includes a wooden walkway and differently shaped obstacles. It was preliminary validated with an able-bodied subject, the same subject walking on commercial prostheses through modified walking boots, and with a prosthetic user. They performed walking firstly on even ground, and secondly on even ground stepping on one of the obstacles. Results in terms of vertical ground reaction force and knee moments in both the sagittal and frontal planes show how the poor performance of commonly used prostheses is exacerbated in case of obstacles. The prosthetic user, indeed, noticeably relies on the sound leg to compensate for the stiff and unstable interaction of the prosthetic limb with the obstacle. Therefore, since the limitations of non-adaptive prosthetic feet in obstacle-dealing emerge from the experiments, as expected, this study justifies the use of the setup for investigating the performance of soft feet on uneven grounds and obstacle negotiation.
\end{abstract}
\protect\copyrightnotice
\section{Introduction}
Approximately half of the lower-limb prosthetic users (LLPUs) report one or more falls per year \cite{kim2022}, which result to be injurious for more than one fourth of them \cite{tobaigy2023}, negatively affecting their quality of life and the burden on the finance of the national health system. 
\par Environmental factors, user-specific (e.g. lack of a proper prosthesis control by the user), and prosthesis-related (e.g. lacking of an appropriate foot-ground clearance, malfunction or failure, other limitations in prostheses performance) factors contribute to the destabilization of the foot support base, leading to falls \cite{kim2019}. Moreover, unevenness often leads to a fall if caught with the prosthesis \cite{kim2022}. Consequently, preventive strategies to mitigate the risk of falling in LLPUs usually include PUs' specific training, environmental modifications, as well as improved prosthetic designs. 
\par Both research and industry have recently attempted to design adaptive prosthetic feet to reduce falls in LLPUs. Some commercial solutions feature an actively controlled ankle range of motion for improved swing ground clearance and stability on all terrains. A few research designs attempt to improve ground adaptability through a 2 Degrees of Freedom (DoFs) ankle joint allowing rotation in both sagittal and frontal planes, or toes dorsiflexion/plantarflexion thanks to a bio-inspired metatarsophalangeal joint \cite{weerakkody2017}. Despite some good adaptive performance on slopes, all these prostheses still embody a flat and stiff foot sole, behaving like a cantilever during the stance phase of walking. Hence, it prevents an effective ground adaptation and obstacle negotiation, jeopardizing LLPUs' stability \cite{weerakkody2017}. 
\par In this scenario, we developed a few years ago a novel soft robotic foot, i.e. the SoftFoot \cite{piazza2016}. Like the human foot filters smaller and bigger unevenness, thanks to the joint work of the longitudinal and transverse arches, and the connective tissue (i.e. plantar fascia) constituting the sole \cite{venkadesan2020}, the SoftFoot features a passive anthropomorphic structure embodying an intrinsic ground adaptability in the sagittal plane. The elastic and flexible sole can adapt to any uneven ground profile, wrapping obstacles, for a compliant interaction with the environment. 
\begin{figure*}[ht]
\abovecaptionskip =0pt
\begin{subfigure}{0.33\textwidth}
\centering
\includegraphics[width=1\linewidth]{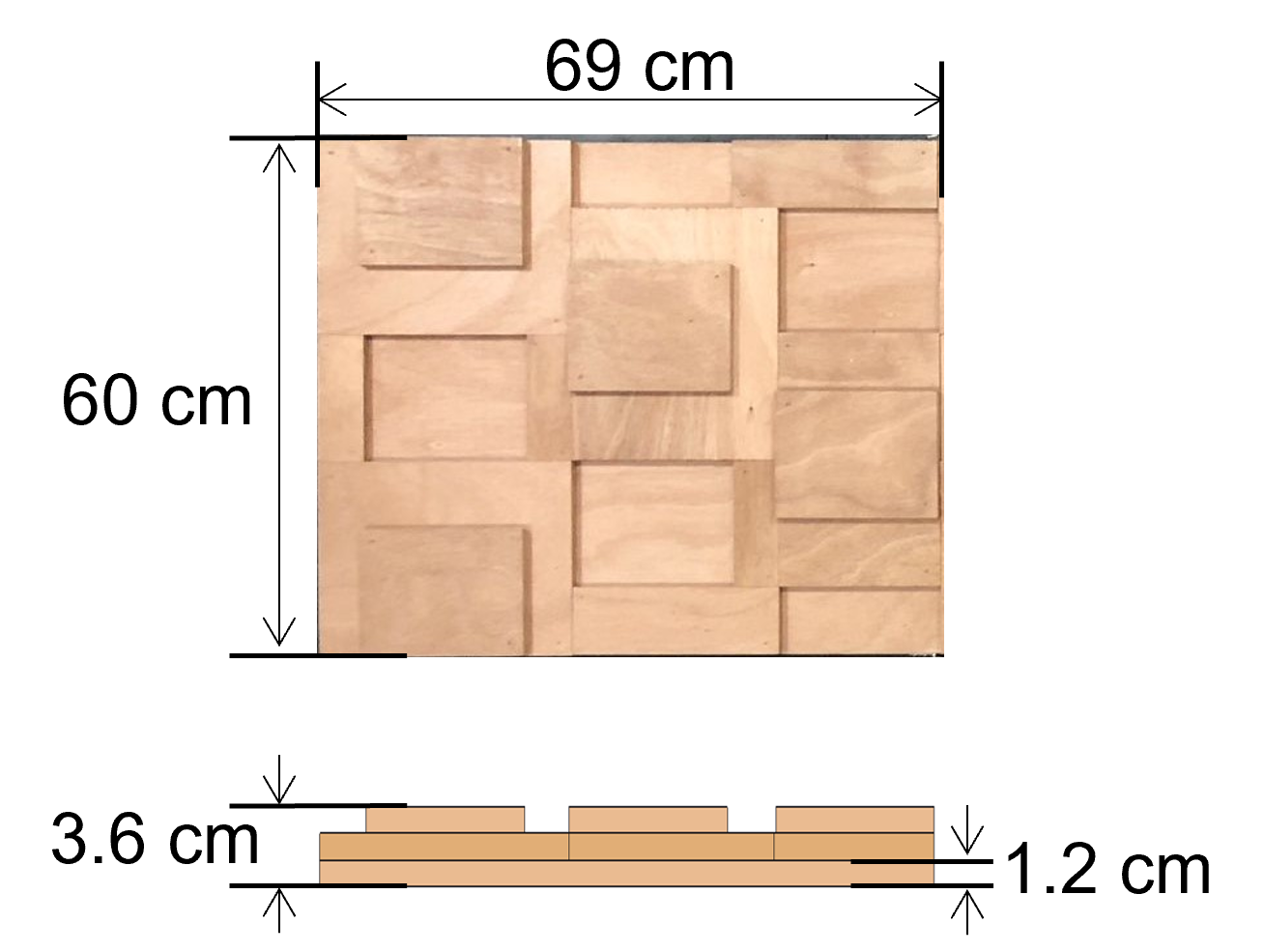} 
\caption{}
\label{uneven}
\end{subfigure}
\begin{subfigure}{0.33\textwidth}
\centering
\includegraphics[width=1.2\linewidth]{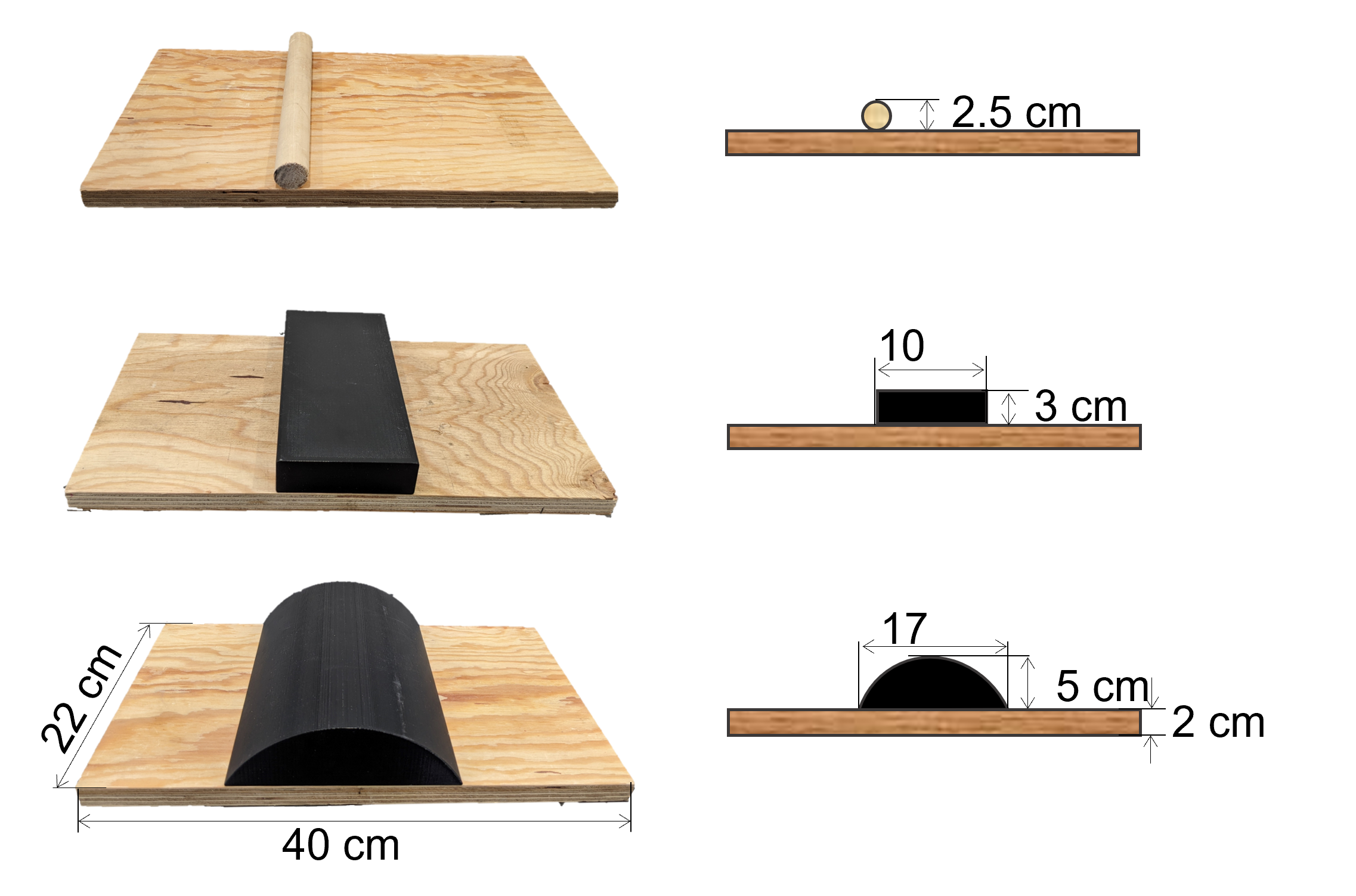}
\caption{}
\label{obstacles}
\end{subfigure}
\begin{subfigure}{0.33\textwidth}
\centering
\includegraphics[width=0.7\linewidth]{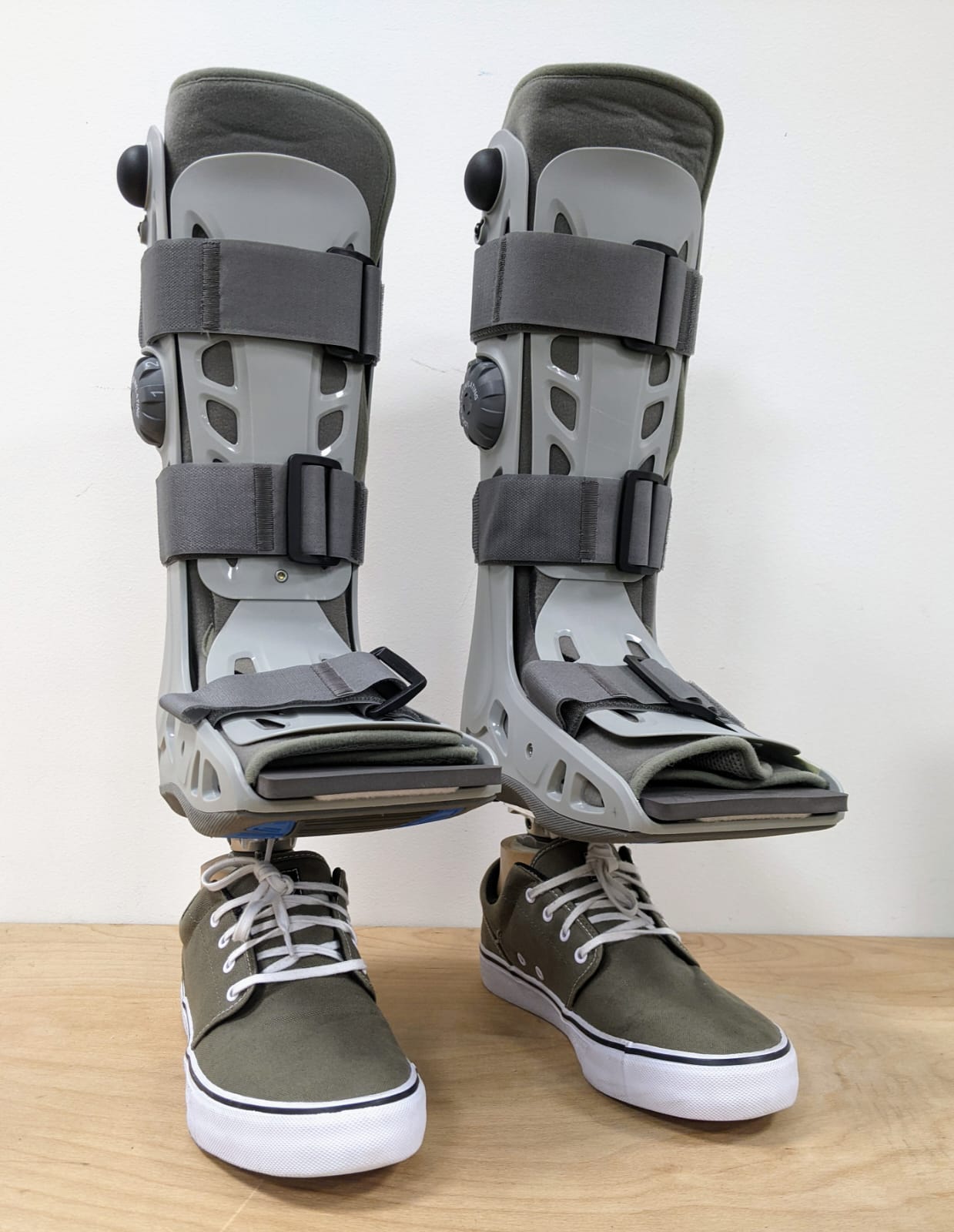}
\caption{}
\label{boots}
\end{subfigure}
\caption{Experimental setup: (a) one of the four 60x69 cm wooden panels built to be placed in a row to form a walkway; (b) differently shaped obstacles, each one screwed on a 40x22x2 cm piece of wood; (c) walking boots modified to assemble two SACH feet (Ottobock, Duderstadt, Germany) with shoes under their sole.}
\label{setup}
\vspace{-0.6cm}
\end{figure*}

\par Despite the good performance achieved during a preliminary benchtop validation \cite{piazza2016}, and when tested on a humanoid robot \cite{catalano2021}, the SoftFoot for prosthetic applications still needs a consistent experimental validation with prosthetic users. The flexible sole may improve stability, reducing the risk of falls, mitigating those gait modifications characterizing walking on unevenness with commercial prostheses, such as reduced speed, modified joint moments, increased metabolic cost, as well as an overall increased mental burden and control effort required from users \cite{kim2022,tobaigy2023,kim2019,chiu2021,mcdonald2021}.
\par To evaluate the performance of soft feet, stressing on their obstacle-dealing capabilities, we have developed a specific experimental setup, which is described in this paper. A standard evaluation method for lower-limb prostheses, indeed, does not exist yet, while benchmarking for performance assessment is an important topic in robotics \cite{torricelli2015}. While tasks such as walking on level grounds, slopes and stairs are usually included in studies on the assessment of the performance of lower-limb prostheses, walking on obstacles or uneven terrains is rarely included \cite{ghillebert2019}, given that commercial prosthetic feet are not primarily designed to effectively adapt to unevenness. 
\par Some studies focus on the biomechanics of LLPUs while stepping over obstacles (e.g. \cite{vrieling2007,buckley2013,deasha2015}). Coleman \textit{et al.} \cite{coleman2016} investigated the changes in able-bodied subjects' biomechanics due to the instability introduced by an uneven walkway, made of wooden blocks assembled on wooden panels in specific positions. Similar designs for a walkway include, for instance, the one described in \cite{curtze2011} to investigate changes in some of the gait parameters of transtibial (TT) prosthetic users wearing commercial prostheses, or the one described in \cite{thomas2020} to study gaze during walking. Some instrumented treadmills were also designed to investigate biomechanical adaptations while walking on uneven grounds over many gait cycles, made of wooden blocks directly attached to the treadmill belt (e.g. \cite{chiu2021,voloshina2018,kent2019}). Nevertheless, most of those setups, by being cumbersome and heavy, limit experimental sessions to the laboratory environment.
\par The setup described here includes a wooden walkway, similar in some aspects to others described earlier, but modular and lightweight, thus portable, and also differently shaped obstacles simulating objects that LLPUs can find on the ground along the way in their daily life, such as branches or stones. It is preliminary validated with an able-bodied subject, walking firstly with a pair of common shoes, and then on commercial rigid prosthetic feet through specifically modified walking boots, and a unilateral TT prosthetic user. The study participants are asked to step on obstacles while walking, rather than stepping over them. In this way, we investigate some of the gait modifications happening in response to unevenness, comparing the performance of some commercial prosthetic feet featuring a flat and stiff sole with the one of an able-bodied subject. We expect the former to be exacerbated by the presence of the obstacle. If that is the case, the described setup is useful to investigate the different performance of adaptive and non-adaptive feet on uneven terrains. Thus, it will be used next for a consistent validation of soft feet. 
\par Setup and experiments are described in Section II. Results are reported in Section III, and discussed in Section IV.
\section{Methods}\label{methods}
\subsection{Participants}
Testing was conducted after obtaining the informed consent from two participants: an able-bodied subject (male, 26 year, 1.87 cm, 75 kg), and a prosthetic user with unilateral osseointegrated TT amputation on his left limb (male, 34 years, 1.90 m, 92 kg) wearing his own carbon fiber prosthetic foot (Triton, Ottobock, Duderstadt, Germany). The study was approved by the Bioethical Committee of the University of Pisa.
\subsection{Experimental setup \& Data collection}
The wooden walkway is made of four 60x69 cm wooden panels with the same design (see Fig. \ref{uneven}) placed in a row, one after the other. The pattern used in each panel is similar in some aspects to the ground profile used in \cite{chiu2021}, which was specifically designed to avoid a 'learning effect' for the user. Each panel is made of wooden rectangles of different sizes, 1.2 cm high, combined in three levels, for an overall height ranging from 1.2 cm to 3.6 cm. Moreover, three differently shaped obstacles were used, similar to those ones used in our previous work described in \cite{piazza2016} (see Fig. \ref{obstacles}). 
\par Kinematic data were collected at 100 Hz using a 12-camera motion capture system (Vicon Motion Systems Ltd., Oxford, UK). Two 60x120 cm force plates (AMTI, Watertown, MA, USA) were used to measure ground reaction forces (GRFs) at 1 kHz. Furthermore, forces and torques at the interface between the prosthetic device and the osseointegrated implant in the participant with TT amputation were measured by a 6 DoFs iPecs load cell (RTC Electronics, Dexter, MI, USA), at a sampling frequency of 1 kHz. It sent data wireless to a receiver connected to the Vicon Lock (i.e. Vicon’s control box for connecting and synchronizing third-party devices with the mocap system), so that this data was synchronized with the other kinematic and kinetic data.
\subsection{Experimental protocol}
\begin{figure}[ht]
\abovecaptionskip = 1pt
\centering
\includegraphics[width=0.48\textwidth]{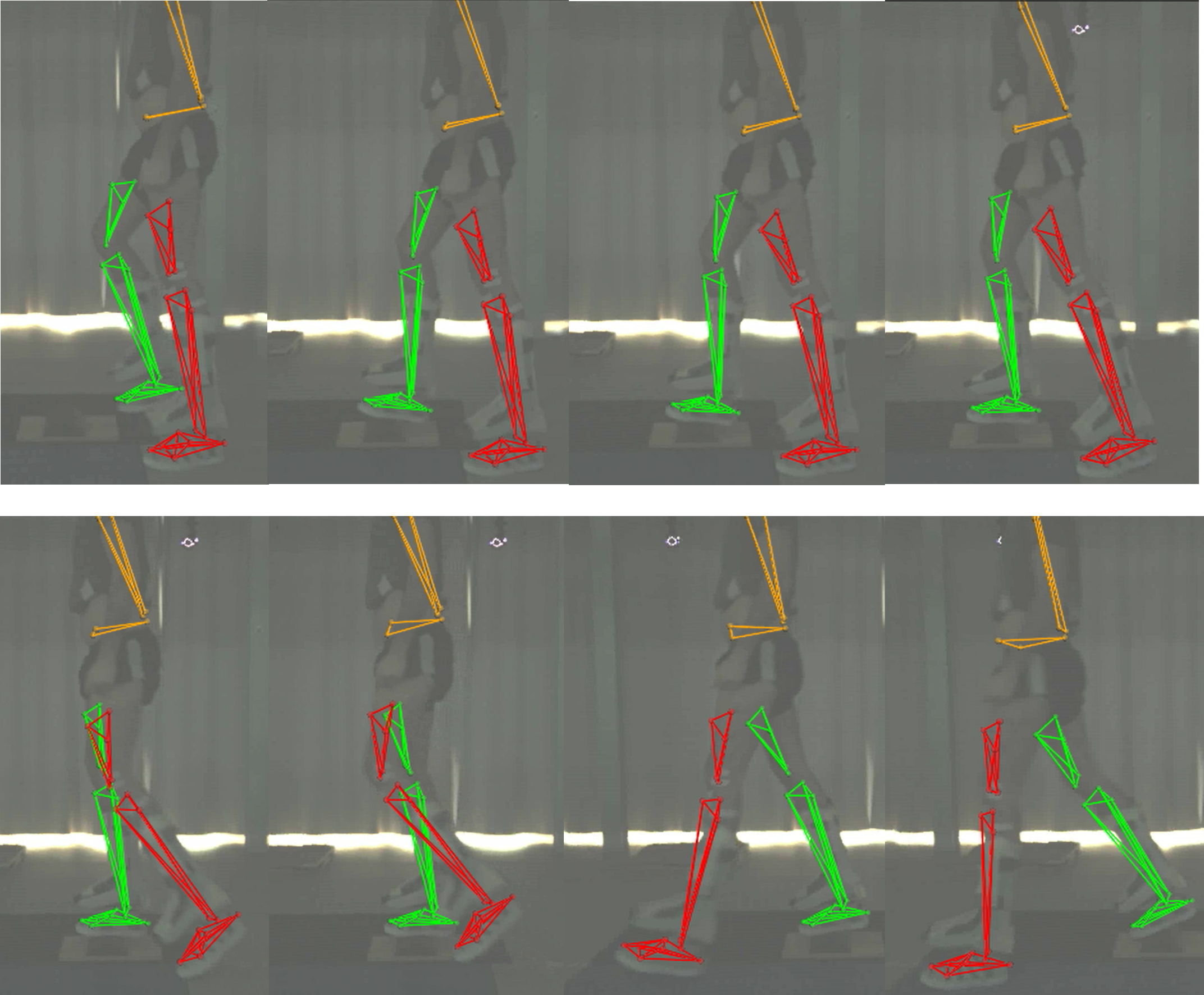} 
\caption{Able-bodied participant wearing modified walking boots with conventional rigid prosthetic feet (SACH by Ottobock). The sequence of the images shows the subject stepping on the arc-shaped obstacle and, in particular, the rigid interaction of the prosthesis with the obstacle profile. Images are obtained as an overlay of 3D marker data onto 2D video in the data capture software Vicon Nexus. Coloured links represent the reconstruction of the body segments based on 3D markers trajectories collected by the Vicon system: green and red respectively for the right and left lower-limbs, orange for the pelvis and trunk.}
\label{stick}
\vspace{-0.6cm}
\end{figure}

In this preliminary validation, the two participants were asked to perform two tasks: walking on level ground at first, and then walking while stepping on the arc-shaped obstacle, placed on one of the force plates embedded in the ground. Being a preliminary validation, we selected the biggest obstacle, i.e. the arc-shaped one, whose height is greater than that of the wooden walkway, and hence it was expected to destabilize the participants' gait the most. For the second task, in order to step on the obstacle, a specific starting point was defined for the two participants, depending on their stride length. They were asked to perform the tasks at self-selected walking speed. Two valid trials were collected for each task. Both participants wore the same pair of skate shoes with flat sole during data collection. The able-bodied subject was instructed to repeat the two tasks wearing first the aforementioned shoes (representing the control), and second a pair of walking boots specifically modified to assemble two conventional prosthetic feet (SACH, Ottobock, Duderstadt, Germany), with shoes, under their sole (see Fig. \ref{boots} and Fig. \ref{stick}). The SACH feet are rigid feet with a flat sole still widely used by LLPUs with a low-level of mobility \cite{delussu2016}. Wearing the boots added a mass and a height of respectively about 4.4 kg and 15 cm to the subject's body mass and height.
\par Kinematics was evaluated from 39 reflective markers placed on the participants' lower limbs and the trunk, according to a modified Cleveland Clinic marker set. The load cell was mounted below the prosthetic implant by a certified prosthetist, and aligned with the other prosthetic components.
\subsection{Data processing}
Data were processed in Matlab (v. 2020b, MathWorks Inc., Natick, MA, USA). Raw kinematic and kinetic data were processed by using a zero-lag, fourth-order, low-pass Butterworth filter, with a cut-off frequency of respectively 6 Hz \cite{gates2012,gates2013} and 20 Hz \cite{grimmer2019,horsak2020}. 
\par A representative trial was selected between the two valid ones collected for each task, based on a visual inspection of the data to identify the one characterized by the most natural walking of the user. Data are reported separately for the leading and the trailing limb. In case of obstacle, the leading limb is always the one stepping on it, and it also coincides with the prosthetic limb in the prosthetic user.
\par The vertical component of the GRFs ($GRF_z$) is displayed as a percentage of the participants' body weight (\%BW) for comparison purposes. The internal knee moments ($M_{KNEE}$) are normalized to the body mass and the shank length (specifically the distance between the knee and the ankle joint), to remove the effect of the height added by the boots. Both variables are also normalized to the stride duration, and resampled to 1500 samples. 
\par GRFs, center of pressure (COP) data and the trajectories of the knee joints were used to estimate the knee joint moments in the sagittal and frontal planes. Those moments are then showed as internal joint torques (due to muscle and joint contact forces) opposing the external torques given by the ground reaction forces \cite{orekhov2019}.
\section{Results} 
\par
\begin{figure}[t] 
\abovecaptionskip = 1pt
    \centering
    \subfloat[]{
    {\includegraphics[trim = 1.39cm 9.5cm 1.2cm 8.5cm, clip, width = 1\linewidth]{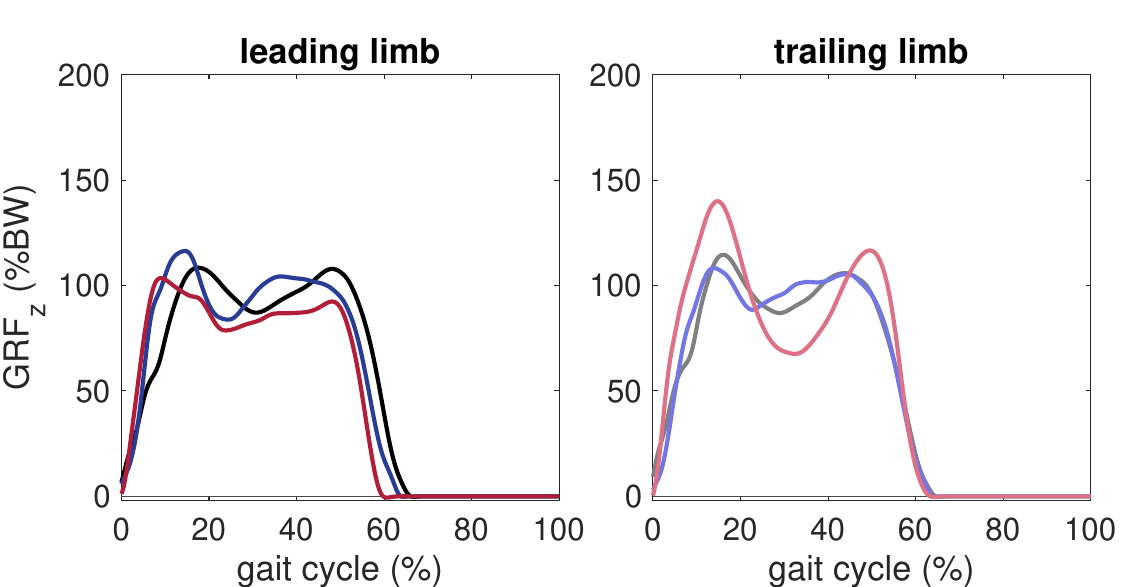}} \label{even}}\\
    \subfloat[]{
    {\includegraphics[trim = 1.39cm 8.8cm 1.2cm 8.5cm,clip, width = 1\linewidth]{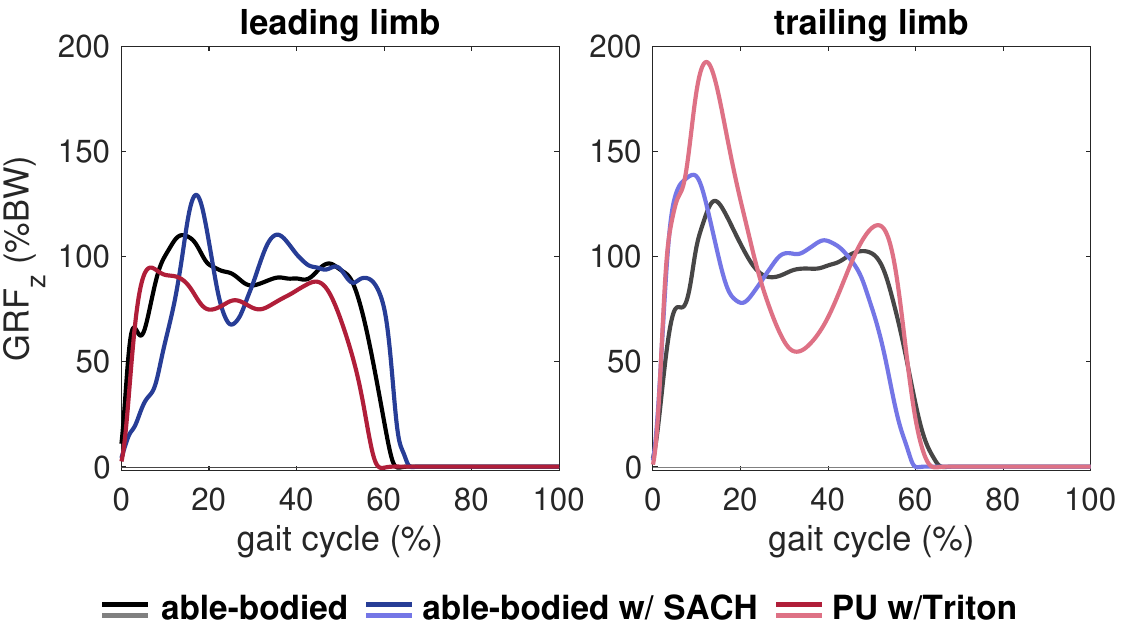}} \label{obs_bigarch}}
\caption{Ground reaction force in the sagittal plane ($GRF_z$) when the participants walk on (a) even ground, and (b) even ground with the arc-shaped obstacle placed on the force plate. On the left side, the leading limb is plotted, while on the right side the trailing limb is plotted. In case of obstacle, the leading limb steps on it. For the prosthetic user (PU) walking with his own carbon fiber prosthetic foot (i.e. the Triton), the leading limb corresponds to his prosthetic limb. Black and gray colors are used for the able-bodied subject, blue shades for the able-bodied subject walking on boots (with SACH feet at the bottom), and red shades for the PU.}
\label{GRFz}
\vspace{-0.6cm}
\end{figure}

 Fig. \ref{GRFz} shows the sagittal component of the GRFs  when the participants walked on even ground with and without the arc-shaped obstacle. 

\par When walking on even ground (see Fig. \ref{even}), the vertical GRF of the control subject is characterized by the typical double-hump profile exceeding body weight, with the first peak ($F_1$) occurring at the end of the loading response phase (about 12\% of the gait cycle), and the second one ($F_2$) towards the end of the stance phase in preparation for swing (about 50\% of the gait cycle). The able-bodied participant wearing boots with the SACH feet shows an increased $F_1$, and a smaller $F_2$ due to the inability of the SACH to provide the required push-off energy. The prosthetic user (PU), on the contrary, exhibits an overall reduced $GRF_z$ between the 10\% and 50\% (i.e. the single-limb support phase) of the gait cycle on his leading limb, corresponding to his prosthetic side, with respect to the able-bodied subject. A larger $F_1$ in the trailing limb of the PU (about 140\%BW) is a consequence of the small $F_2$ measured in his leading limb, which is again caused by the lack of a proper push-off in the prosthetic limb. 
\par This trend is exacerbated when stepping on the obstacle (see Fig. \ref{obs_bigarch}). The able-bodied subject walking on rigid prosthetic feet (see Fig. \ref{stick}) presents the largest $F_1$. The PU using a carbon fiber foot exhibits again a reduced $GRF_z$ during the single-limb support phase of the prosthetic limb, due to his tendency to load more the sound leg, as the larger values of $GRF_z$ in his trailing limb demonstrate. Specifically, the very large $F_1$ recorded at the PU's sound leg compensates for the lack of adaptability to the ground and the reduced push-off power generation in the prosthesis. The same trend characterises also the $GRF_z$ measured at the trailing limb of the able-bodied subject with boots (see the corresponding $F_1$ peak in Fig. \ref{obs_bigarch}) after stepping on the obstacle.
\par
\begin{figure*}[h]
\abovecaptionskip = 1pt
\centering
\subfloat[]{
{\includegraphics[trim = 1.6cm 9.45cm 1cm 9cm, clip, width = 0.49\linewidth]{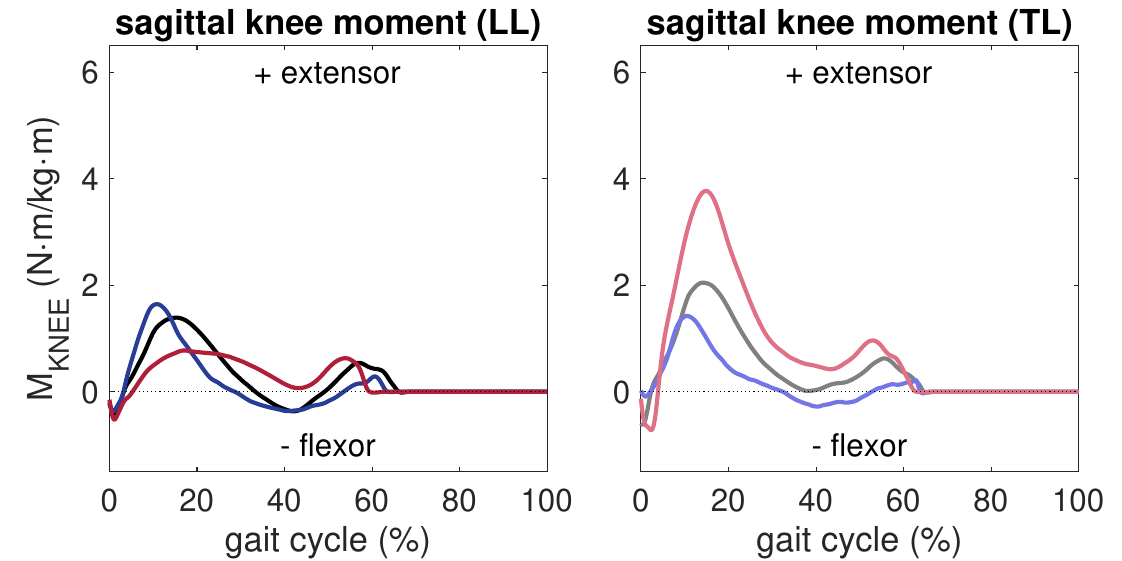}}\label{kneeT_even_sagittal}}
\subfloat[]{
{\includegraphics[trim = 1.39cm 9.45cm 1cm 9cm, clip, width=0.49\linewidth]{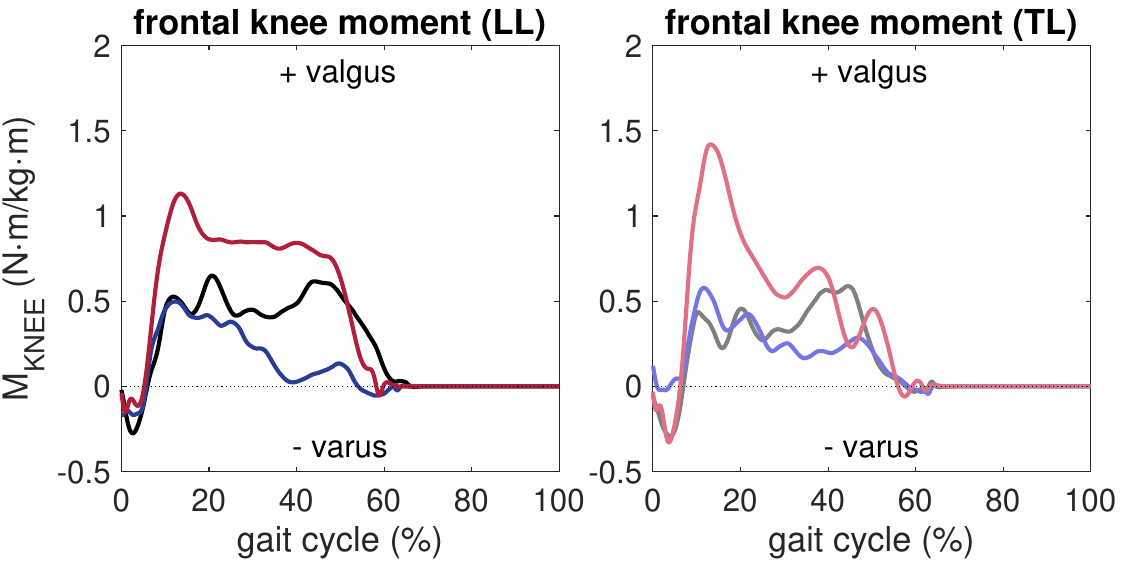}}
\label{kneeT_even_frontal}}\\
\subfloat[]{
{\includegraphics[trim = 1.6cm 9.45cm 1cm 9cm, clip, width=0.49\linewidth]{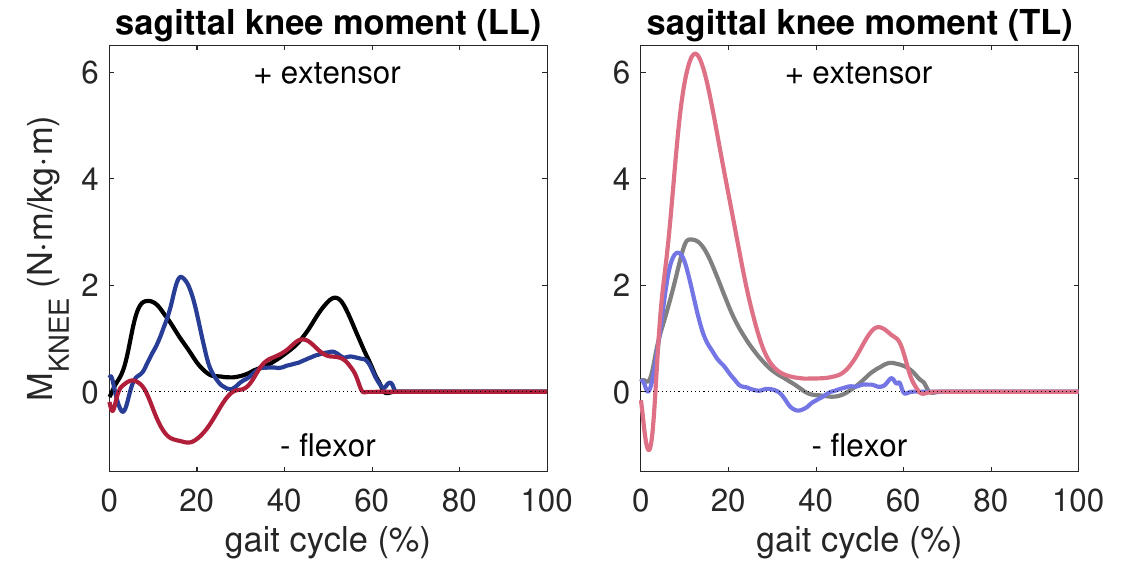}}
\label{kneeT_bigarch_sagittal}}
\subfloat[]{
{\includegraphics[trim = 1.39cm 9.45cm 1cm 9cm, clip, width=0.49\linewidth]{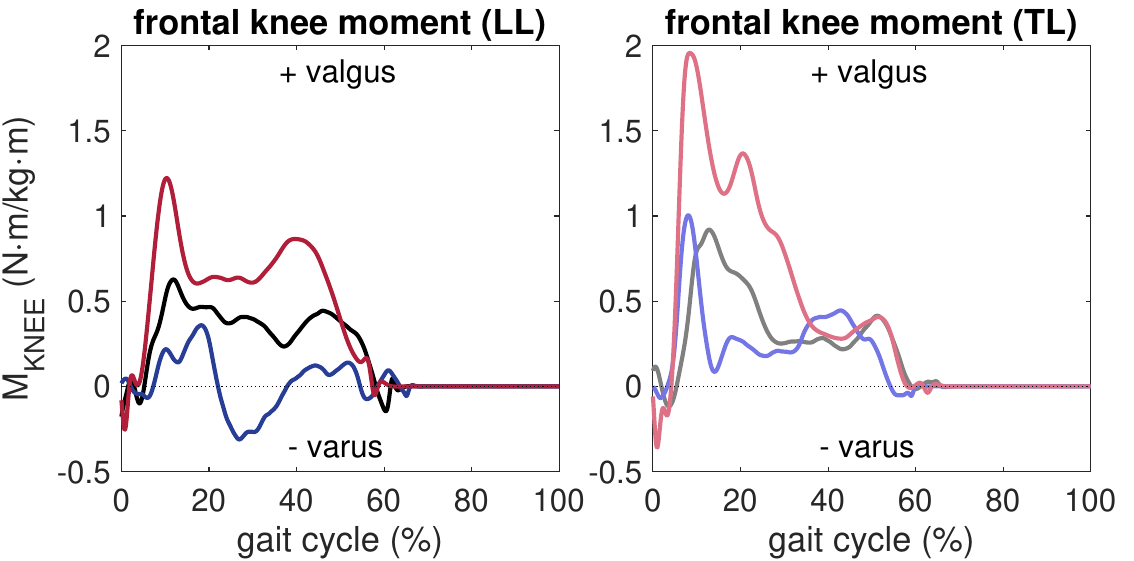}}
\label{kneeT_bigarch_frontal}}\\
\subfloat{
{\includegraphics[width=0.43\linewidth]{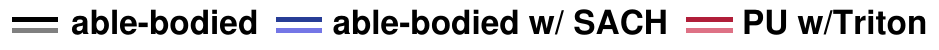}} \label{legend}}\\
\caption{Internal knee moments ($M_{KNEE}$) in the leading limb when the participants walk on (a,b) even ground, and (c,d) even ground with the arc-shaped obstacle placed on the force plate. The knee moment in the sagittal plane is displayed in the two columns on the left side, for the leading (LL) and the trailing (TL) limb, while the moment in the frontal plane is showed in the two columns on the right side for the two limbs. The participants always stepped on the obstacle with their leading limb, corresponding to the prosthetic limb for the PU. Black and gray colors are used for the able-bodied subject, blue shades for the able-bodied subject walking on boots (with SACH feet at the bottom), and red shades for the PU.}
\label{kneeT}
\vspace{-0.6cm}
\end{figure*}

 Fig. \ref{kneeT} shows the internal moments at the ankle joint in the sagittal and frontal planes, when participants walked on the even ground without (Fig. \ref{kneeT_even_sagittal}, \ref{kneeT_even_frontal}) and with the arc-shaped obstacle (Fig. \ref{kneeT_bigarch_sagittal}, \ref{kneeT_bigarch_frontal}). A typical double-hump profile characterizes the control participant's knee sagittal moment during level walking (see Fig. \ref{kneeT_even_sagittal}). It corresponds to internal extensor moments ensuring stability during knee flexion at the beginning (about 10\%) and before the end of stance. However, the PU exhibits smaller values of the first extensor knee moment in the sagittal plane, due to reduced knee flexion angles reported in the first half of stance in LLPU with transtibial amputation. 
\par In case of  obstacle (Fig. \ref{kneeT_bigarch_sagittal}), the control participant shows a greater extensor sagittal moment in the leading limb throughout stance to oppose the larger knee flexion occurring to step on the obstacle. A similar behaviour can be seen in case of walking with boots. Differently, a flexor sagittal moment during the first half of the stance phase characterises PU's walking on obstacle. This is caused by the extended knee joint used to step on the obstacle.
\par Furthermore, participants show a sagittal moment with a high positive peak in their trailing limb. It occurs during the transition from loading response to mid-stance in response to obstacle negotiation in the leading limb, to regain stability after the instability introduced by the obstacle, loading the trailing limb. This is more evident in the PU, in which the first peak value of the internal sagittal knee moment in the sound limb is already greater than the control subject during level walking, because of the poor performance of the prosthetic leg. It then almost triples the control participant in case of obstacle.
\par In the frontal plane (Fig. \ref{kneeT_even_frontal}, \ref{kneeT_bigarch_frontal}), the increase of the stance valgus moment in the PU is also noticeable: it almost doubles the control participant in the leading limb, while it becomes even higher than that in the trailing limb.


\section{Discussion} \label{discussion}
Many studies reported in literature investigated the biomechanics of transtibial prosthetic users during gait, comparing it with that of able-bodied subjects, and showed the poor performance of commercial prosthetic feet. These results worsen in case of walking on unstructured environments \cite{kent2019,gates2012}. The results of this preliminary validation of the experimental setup are in agreement with those findings, since they highlight the bad performance of prosthetic feet currently on the market featuring a stiff sole, with respect to able-bodied subjects.
\par In Fig. \ref{GRFz}, for instance, the limited capabilities of commercial feet in obstacle negotiation stand out. The SACH foot, when worn by the able-bodied subject through the boots, produces a higher $GRF_z$ peak during the transition from loading response to mid-stance when stepping on the obstacle, likely due to the rigid impact of the foot on it (see also Fig. \ref{stick}). The PU always tends to make contact with the obstacle by using the central area of the prosthetic foot sole (from analysing some videos of the trials), to avoid high impact forces in case of a heel strike directly on the obstacle. Also, PUs usually rely on their sound leg while walking, loading it mostly \cite{kovac2009}, to ensure stability, as a $GRF_z$ in the prosthetic limb with peaks smaller than 100\%BW demonstrates. This results in a large $GRF_z$ in the PU's trailing limb, i.e. the sound leg, during single-limb support, with a typical higher $F_1$ caused by the lack of a proper power generation during push-off for forward body propulsion in the prosthetic limb (small $F_2$, in line with literature, e.g. \cite{amma2021}). This first peak in the sound limb becomes very high in case of obstacles, compensating for the poor performance of the rigid and non-adaptive foot-obstacle interaction. A similar trend can be noticed also in the able-bodied participant walking on SACH feet, although the results obtained in this last case need to be cautiously compared to the PU and the control, because of the limitations introduced by the use of boots.
\par In addition, the internal flexor sagittal moment measured in the first half of stance at the prosthetic limb in case of obstacle reflects the lack of ground adaptability of the prosthesis. The PU, indeed, tends to step on the obstacle keeping his knee extended, pivoting on the contact point between the rigid prosthetic foot sole and the obstacle. A similar strategy certainly challenges PU's stability: placing the prosthesis in the wrong way on the obstacle or any additional destabilizing force acting on the user might lead to a fall. Also the very high extensor sagittal moment occurring in the PU's sound limb after going beyond the obstacle is a consequence of the rigid and unstable interaction of the prosthesis with the obstacle profile, which leads to a very high load and large knee flexion angles during loading response in the sound limb.
\par Therefore, the experimental setup described in this paper allowed us to discern the poor performance of an able-bodied subject walking on SACH feet and, above all, of a TT PU with respect to the one of an able-bodied subject. The stepping-on-the-obstacle task, in particular, allowed to highlight the limitations and the drawbacks of the use of a prosthetic foot with a flat and rigid sole, differently from most of the studies focusing on obstacle avoidance. 
\par As a consequence, this preliminary validation of the setup justifies its future use to test our soft feet with LLPUs, through a consistent experimental campaign and with an appropriate number of participants to get statistically significant results. We hypothesize that, when wearing a soft foot, the prosthetic users will be able to rely more on their prosthetic limb, partially unloading the sound side, thanks to a compliant interaction with the ground, which will no longer be a source of instability. This means $GRF_z$ larger than BW in the prosthetic limb, without high first peaks due to a rigid impact with unevenness, and with smaller peaks in the sound side. As a consequence, soft feet will contribute to reduce those gait modifications and compensatory strategies arisen from the use of commercial non-adaptive feet, e.g. the knee moments will be closer to those seen in the able-bodied. 
\par The main limitations of the presented study are the small number of trials collected, and the results reporting only data from one representative step for each limb. Another limitation is the small number of tasks performed. However, future testing of soft feet will be conducted using the whole setup described in Section II (i.e. the three differently shaped obstacles, and the uneven walkway), and collecting a significant number of valid trials (at least three) for each limb and task, reporting mean values and SD in the study results. Forces and moments from the iPecs load cell will provide valuable kinetic information over more than one stride, and not restricted to the laboratory environment. Also, the use of a setup that is low-cost, lightweight, and easily portable will allow to overcome the limitations of instrumented treadmills or long single-piece walkways. Lastly, the use of single obstacles with different shapes will also allow in future studies to better characterize the behaviour of soft feet sole when LLPUs step on them, mapping for instance plantar pressure. For this purpose, more than a walkway, obstacles will ensure repeatability and reproducibility of the experiments.

\section{Conclusions}
In this study we describe the experimental setup we built to compare the performance of soft and rigid prosthetic feet during walking on uneven grounds. We recently developed, indeed, a soft robotic foot for enhanced adaptability on uneven grounds, whose performance still needs to be assessed on lower-limb prosthetic users. The setup consists of four wooden panels forming a walkway, and three differently shaped obstacles. We preliminary validated it with an able-bodied subject, as a control, the same subject wearing walking boots with conventional rigid prosthetic feet, and a transtibial unilateral prosthetic user wearing his own carbon fiber foot. We asked them to walk on level ground first, and then on the same ground but stepping on an arc-shaped obstacle. 
\par Results obtained confirmed the poor performance of commercial prosthetic feet featuring a stiff sole when dealing with obstacle, placing at risk the prosthetic user's stability, which might result in a fall. This is in line with previous findings about lower-limb prosthetic users' gait. 
\par Therefore, this low-cost and portable experimental setup can be used to shed light on the adaptive capabilities of soft feet through a consistent experimental campaign with prosthetic users. The benefits of a compliant interaction with the ground may greatly improve the quality of life of people living with a lower-limb amputation.

\section*{Acknowledgment}
The authors thank M. Barbarossa, G. Rosato, and E. Sessa for their valuable help in the manufacturing of the setup, and the execution of the preliminary experimental testing on the able-bodied subject and the prosthetic user. 

\bibliographystyle{IEEEtran}
\bibliography{biblio_ICORR23}

\end{document}